\documentclass[journal]{IEEEtran}
\usepackage{amsmath}
\usepackage{amssymb}

\usepackage{bbding}
\usepackage{amsmath}
\usepackage{amssymb}
\usepackage{array}
\usepackage{booktabs}
\usepackage{multirow}
\usepackage{multirow}
\usepackage{setspace}
\usepackage{color,subfigure}
\usepackage{mathrsfs}
\usepackage{graphics}
\usepackage{epsfig}
\usepackage{epstopdf}
\usepackage{url}
\usepackage[square,numbers,sort&compress]{natbib}
\usepackage{amsmath,bm}
\usepackage{stfloats}
\usepackage[pagebackref=false,breaklinks=true,letterpaper=true,colorlinks,bookmarks=false]{hyperref}

\ifCLASSINFOpdf

\else

\fi

\begin{document}
%
\title{AIParsing: Anchor-free Instance-level \\ Human Parsing}
%
%
%

\author{Sanyi Zhang,~\IEEEmembership{Member,~IEEE,}
        Xiaochun Cao,~\IEEEmembership{Senior Member,~IEEE,}
        Guo-Jun Qi,~\IEEEmembership{Fellow,~IEEE,}\\
        Zhanjie Song,~\IEEEmembership{Member,~IEEE,}
        and~Jie~Zhou,~\IEEEmembership{Senior Member,~IEEE}
\thanks{
This work is supported in part by the Shenzhen sustainable development project (No. KCXFZ20201221173013036), in part by National Natural Science Foundation of China (No. 62132006, U2001202), and in part by Open Project Program of State Key Laboratory of Virtual Reality Technology and Systems, Beihang University (No.VRLAB2021C06)
(Corresponding author: Xiaochun Cao.)

S. Zhang is with the State Key Laboratory of Information
Security, Institute of Information Engineering, Chinese Academy of
Sciences, Beijing 100093, China, and also with Georgia Tech Shenzhen Institute, Tianjin University, Shenzhen, 518055, China (e-mail: zhangsanyi@iie.ac.cn).

X. Cao is with School of Cyber Science and Technology, Shenzhen Campus, Sun Yat-sen University, 518107, China (e-mail: caoxiaochun@mail.sysu.edu.cn).

G.-J. Qi is with Futurewei Technologies, Seattle, WA, 98006, USA (e-mail: guojunq@gmail.com).

Z. Song is with Georgia Tech Shenzhen Institute, Tianjin University, Shenzhen, 518055, China
(e-mail: zhanjiesong@tju.edu.cn).

J. Zhou is with the Department of Automation, Tsinghua
University, Beijing 100084, China (e-mail:jzhou@tsinghua.edu.cn).
}
}
%
%

\markboth{}%
{Shell \MakeLowercase{\textit{et al.}}: Bare Demo of IEEEtran.cls for IEEE Journals}
%



\maketitle

\begin{abstract}
Most state-of-the-art instance-level human parsing models adopt two-stage anchor-based detectors and, therefore, cannot avoid the heuristic anchor box design and the lack of analysis on a pixel level. To address these two issues, we have designed an instance-level human parsing network which is anchor-free and solvable on a pixel level. It consists of two simple sub-networks: an anchor-free detection head for bounding box predictions and an edge-guided parsing head for human segmentation. The anchor-free detector head inherits the pixel-like merits and effectively avoids the sensitivity of hyper-parameters as proved in object detection applications. By introducing the part-aware boundary clue, the edge-guided parsing head is capable to distinguish adjacent human parts from among each other up to 58 parts in a single human instance, even overlapping instances. Meanwhile, a refinement head integrating box-level score and part-level parsing quality is exploited to improve the quality of the parsing results.
Experiments on two multiple human parsing datasets (\textit{i.e.}, CIHP and LV-MHP-v2.0) and one video instance-level human parsing dataset (\textit{i.e.}, VIP) show that our method achieves the best global-level and instance-level performance over state-of-the-art one-stage top-down alternatives.

\end{abstract}


%

\section{Introduction}
\IEEEPARstart{I}{nstance-level} human parsing is a fundamental challenge in the fields of computer vision and multimedia, which focuses on pixel-wise human-centric analysis in the world. It aims to not only distinguish various human instances but also segment each instance into accurate human parts. Accordingly, instance-level human parsing can provide human details of each instance, which plays an active role in human-related tasks, such as dense pose estimation \cite{alp2018densepose, yang2019parsing}, person re-identification \cite{li2018harmonious, miao2019pose}, human-object interaction \cite{chao2015hico, gkioxari2018detecting, qi2018learning, zhou2021cascaded}, visual try-on \cite{hsieh2019fashionon, dong2019towards}, fashion editing \cite{dong2020fashion}, human part analysis \cite{yang2020hier}, fashion landmark detection \cite{wang2018attentive}, human-centric video analysis
in complex events \cite{lin2020human}, and so on.

\begin{figure}[t]
  \centering
  \includegraphics[width=0.9\linewidth]{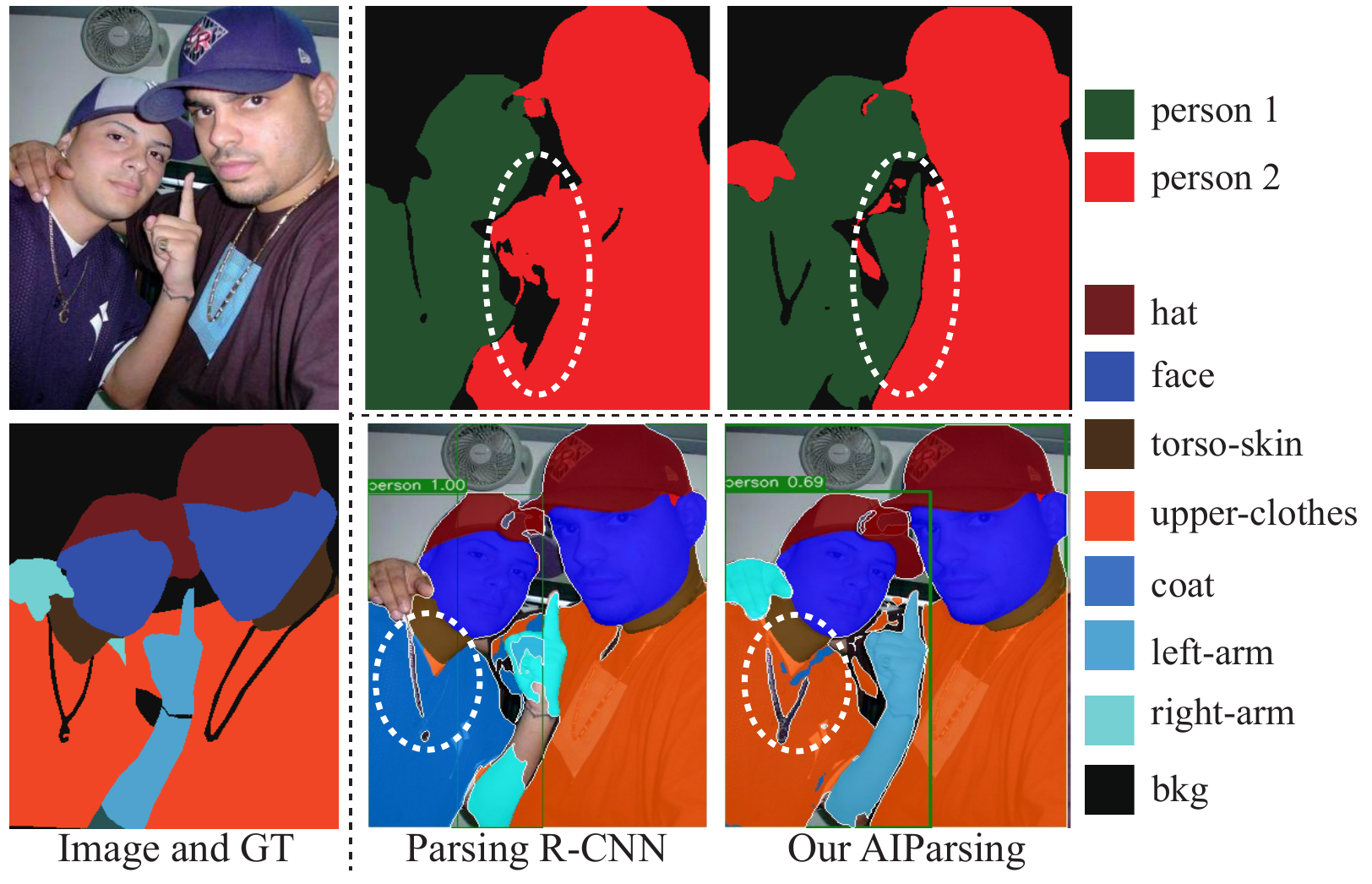}
  \caption{ The motivation of the proposed AIParsing.
   From the two indicated regions (with white color), our proposed AIParsing can distinguish different instance regions (first row) and boundaries among adjacent human parts (second row) than Parsing R-CNN \cite{yang2019parsing} while equipped with the same ResNet-50 backbone.  }
  \label{fig:first}
\end{figure}

Recently, Mask R-CNN \cite{he2017mask} and its improvements \cite{lee2020centermask, chen2019hybrid, huang2019mask, cheng2020boundary} have made great progress in instance segmentation tasks. Following the same spirit, current state-of-the-arts on instance-level human parsing still follow the Mask R-CNN \cite{he2017mask} model. Specifically, the end-to-end framework of the RP R-CNN \cite{yang2020eccv} which utilizes a two-stage human instance detector (\textit{i.e.}, Faster R-CNN \cite{ren2015faster}) to extract anchor-based region proposals and conduct fine-grained human part parsing.

Inevitably, the current two-stage anchor-based human detector has two limitations. First, the performance of the two-stage object detector is sensitive to hyper-parameters pre-defined in the anchor generation \cite{ren2015faster, lin2017focal}, \textit{e.g.}, aspect ratio, anchor area, and scale, \textit{etc}. We need to carefully tune or re-design anchor boxes for new datasets or new detection tasks. Furthermore, choosing suitable bounding box samples with a high recall rate and controlling the imbalance rate of positive and negative samples cause extra computational and memory costs. Second, the two-stage anchor-based detector is not per-pixel prediction oriented, which limits the consistency of the per-pixel parsing module in the human instance parsing task. Thus, exploring a fully convolutional solvable, anchor-free and one-stage detector that avoids hyper-parameter sensitivity meets the need of human instance analysis tasks.
To segment human instances with fine-grained categories, the fully convolutional networks based detector is beneficial to incorporate detection and parsing into a unified framework. Thus, FCOS \cite{tian2019fcos} is introduced to detect various human instances because of its per-pixel comprehensive representation and satisfactory human detection results.

Besides, for the instance-level human parsing task, only focusing on a powerful and elegant FCN-wise human detector is insufficient since obtaining accurate instance parsing regions with fine-grained categories is the ultimate goal. Based on this purpose,
the edge information, which provides a useful cue for distinguishing the boundaries of adjacent human parts from a single human instance or different instances, is provided by a simple edge prediction branch in parallel with the human parsing branch. Therefore, the instance detector ideally provides accurate bounding box results (see Fig. \ref{fig:first} second row, our method provides relatively correct regions for the right human instance), and the human instance parsing module can focus on the fine-grained classification. As shown in Fig. \ref{fig:first}, incorporating edge clues predicts accurate instance regions even though instances are in occlusion, and it is also successful at separating the adjacent human part regions.

What's more, since instance-level human parsing task not only tackles instance-level detection but also handles part-level parsing, it is near impossible to obtain ideal human parsing results merely using pixel-level cross-entropy loss for optimization.
Incorporating instance-level and part-level information provides complementary aspects to enhance the parsing results.
For the instance-level aspect, the box-level score, which aims at choosing high-quality predicted bounding boxes through sorting box scores, is useful for improving the instance-level metric scores by filtering out low-quality predicted bounding boxes. For the part-level aspect, since part-level parsing quality is also important where $\text{AP}^{\text{p}}$ or IoU metrics care, exploring the structure-aware or part-aware criteria, \textit{e.g.}, IoU loss, is useful to refine the parsing results. With these considerations, a refinement head fusing these two conducive clues is constructed to improve the human parsing results with higher global-level and part-level metric scores.

Consequently, a novel one-stage \textbf{A}nchor-free \textbf{I}nstance-level human \textbf{Parsing} network (\textbf{AIParsing}) is proposed, the network embodies an anchor-free human detector (FCOS) and a novel edge-guided parsing branch to generate instance-level human parsing results. Additionally, a comprehensive metric loss function is employed that covering hierarchical enforcements from pixel-level to instance-level. The proposed AIParsing achieves state-of-the-art global metric score (mIoU) and instance-level metric scores ($\text{AP}_{\text{vol}}^{\text{p}}$, $\text{AP}_{\text{50}}^{\text{p}}$ and $\text{PCP}_{\text{50}}$) on two image multiple human parsing datasets, \textit{i.e.}, CIHP  \cite{gong2018instance} and LV-MHP-v2.0 \cite{zhao2018understanding}. In addition, the proposed AIParsing also achieves the state-of-the-art performance of 52.2\% ($\text{AP}_{\text{vol}}^{\text{r}}$) on video instance-level human parsing dataset, \textit{i.e.}, VIP \cite{zhou2018adaptive}.
The main contributions are summarized as follows:

\begin{itemize}
\item[-]A FCN-like anchor-free detector is employed to improve the current RPN-based detector, which leads to a FCN-solvable framework fitting the ultimate pixel-wise instance-level parsing task.
\item[-]An edge-guided parsing module is proposed to accurately parse human regions, which exploits edge clues to distinguish the boundaries of adjacent human parts within the instance and overlapping instance regions. Two complementary clues, \textit{i.e.}, mIoU-score and mIoU-loss, are fused to further improve the quality of the parsing results through the box-level score and part-level parsing quality, respectively.
\item[-]The proposed AIParsing achieves state-of-the-art parsing results on two popular multi-human parsing image benchmarks (\textit{i.e.,} CIHP \cite{gong2018instance} and LV-MHP-v2.0 \cite{zhao2018understanding}) and one video instance-level human parsing benchmark (\textit{i.e.}, VIP \cite{zhou2018adaptive}).
    The proposed AIParsing can also be served as a baseline towards solving instance-level parsing tasks due to the per-pixel nature.
\end{itemize}

The remaining parts of our paper are as follows. In section \ref{relatework}, we introduce the related work. Then, the details of the methodology are presented in Section \ref{method}. Experiments on two instance-level human parsing image benchmarks and one multi-human parsing video benchmark are discussed in Section \ref{experiment}.
Finally, we conclude in Section \ref{conlusion}.

\section{Related Work}\label{relatework}

In this section, we firstly introduce current anchor-free and anchor-based object detectors, then we will discuss instance-level human parsing related solutions.

\subsection{Anchor-based vs. Anchor-free Object Detector}

In the object detection task, there are two mainstream detectors according to the bounding box selection, anchor-based and anchor-free detectors.
Generally, anchor-based object detectors generates a series of candidate proposals according to pre-defined anchor sizes first and then performs object classification and bounding-box offset regression on these proposals. Recently, Region Proposal Network (RPN) based anchors have been certificated their high accuracy and effectiveness on the object detection task, such as Faster R-CNN \cite{ren2015faster}, SSD \cite{liu2016ssd}, YOLOv2 \cite{redmon2017yolo9000}, and YOLOv3 \cite{redmon2018yolov3}.
However, the complex hyper-parameters of anchor-based proposals restrict the generalization capability, we need to carefully design and tune the hyper-parameters for new detection tasks. Moreover, the sampling of anchor boxes also requires complicated computation, such as calculating the Intersection over Union (IoU) with ground-truth boxes.
Although these anchor-based methods provide powerful detection accuracy, exploring anchor-free or proposal-free object detectors without heuristic tuning will enhance the generalization ability. Besides, as with other dense prediction tasks, solving object detection in the way of per-pixel prediction will make these vision tasks formulate in a unified framework.
Thus, anchor-free or proposal-free detectors are proposed to solve these two problems, which utilize points to predict bounding boxes and make full use of fully convolutional networks, such as DenseBox \cite{huang2015densebox}, YOLOv1 \cite{redmon2016you}, CornetNet \cite{law2018cornernet}, FCOS \cite{tian2019fcos}, ExtremeNet \cite{zhou2019bottom}, and CenterNet \cite{zhou2019objects}. YOLOv1 \cite{redmon2016you} predicts bounding boxes at points near the center of objects. CornerNet \cite{law2018cornernet} detects a pair of corners (top-left and bottom-right) of a bounding box and groups them to form the final detected bounding box. In particular, FCOS \cite{tian2019fcos} is an anchor-free per-pixel detection method, which predicts a 4-D vector and a class label to represent a bounding box. The 4-D vector is the distance from the location to the four sides of the bounding box.
And Anchor-free based detectors are also explored in instance segmentation tasks, like \cite{chensupervised2020, lee2020centermask, wu2020improved, chen2020blendmask}. The video instance segmentation task \cite{wang2021end, li2020delving} is also inspired by the anchor-free detection for predicting accurate detection and segmentation of the target frame.

\begin{figure}[t]
  \centering
  \includegraphics[width=0.8\linewidth]{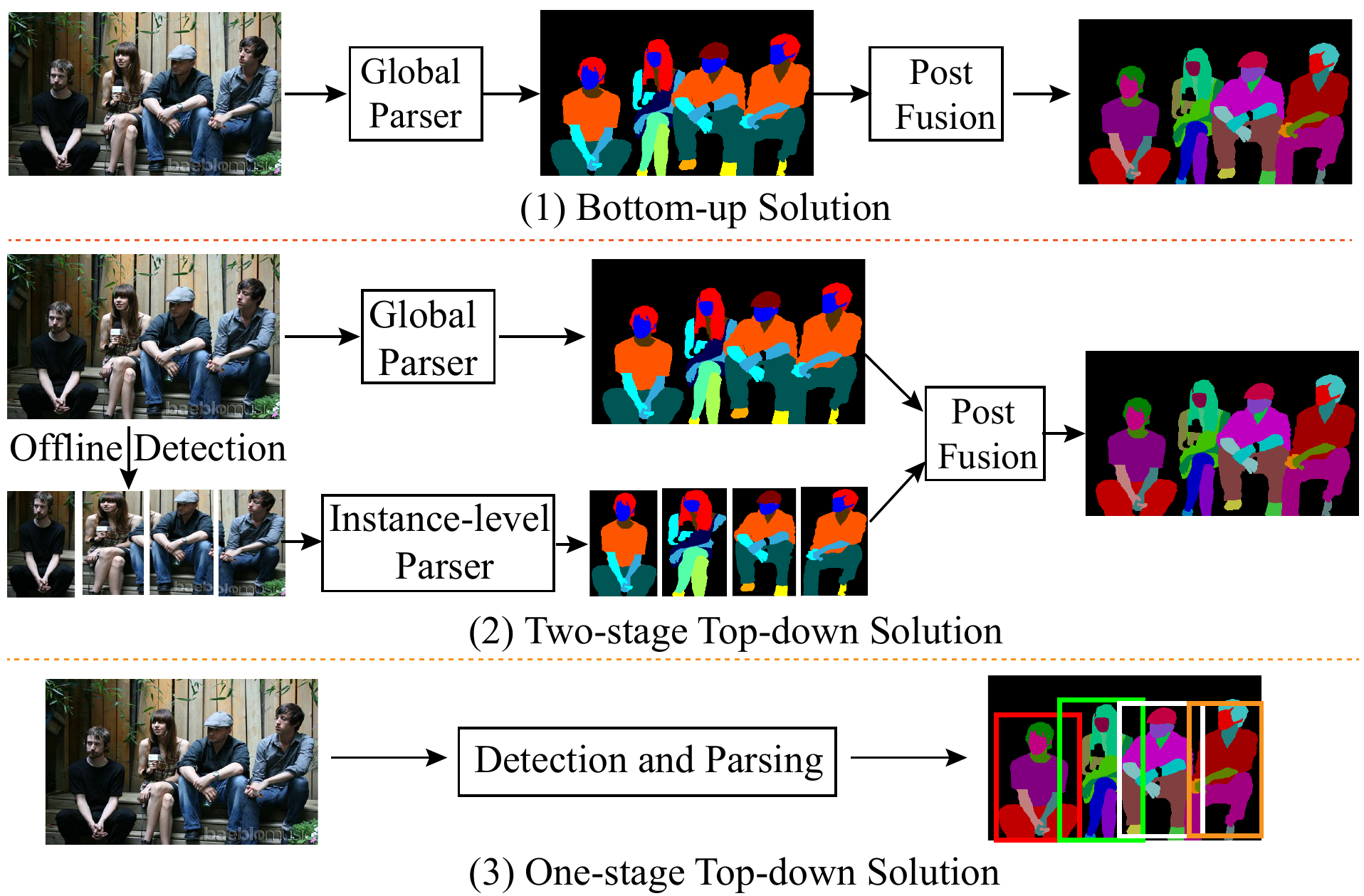}
  \caption{
    Different instance-level human parsing solutions.
   }
  \label{fig:pipeline}
\end{figure}

\subsection{Instance-level Human Parsing}
Current solutions for instance-level human parsing can be divided into three types: bottom-up, two-stage top-down, and one-stage top-down.

\noindent\textbf{Bottom-up}: Bottom-up related solutions \cite{he2019grapy, gong2019graphonomy, gong2018instance} view the instance-level human parsing task as a global fine-grained semantic segmentation task. They mainly focus on firstly predicting accurate
human parsing results on a pixel-level (global parser), and then grouping pixels into different instances according to pre-designed cues (post fusion), as shown in Fig. \ref{fig:pipeline} (1). Gong \textit{et al.} \cite{gong2018instance} proposes a detection-free bottom-up based Part Grouping Network (PGN) for multi-human parsing. PGN generates human instance results by utilizing extra edge information.
Moreover, Graphonomy \cite{gong2019graphonomy} and Gray-ML \cite{he2019grapy} explore the underlying label semantic relations in a hierarchical graph structure.
These methods can achieve satisfactory results on global human parsing, but the instance-level performance is relatively weak since it lacks the human detection branch. Recently, Zhou \textit{et.al.} \cite{zhou2021multi} proposes a novel bottom-up regime (MG-Parsing) that fuses category-level human parsing and multiple human pose estimation into an end-to-end differentiable way. Specifically, a dense-to-sparse projection field connects the dense human parsing annotation with the sparse keypoints, which creates a multi-granularity human representation learning for the instance-level human parsing task.

\noindent\textbf{Two-stage top-down}: Unlike bottom-up based methods, two-stage top-down based methods \cite{ruan2019devil, liu2019braidnet, ji2019learning, zhang2020human} firstly adopt a well-trained object detector \cite{girshick2018detectron} to detect human instances, then the detected instances are utilized to obtain human instance parsing results, as shown in Fig. \ref{fig:pipeline} (2). Ruan \textit{et al.} \cite{ruan2019devil} proposes a Context Embedding with Edge Perceiving (CE2P) framework for the single human parsing task. Ruan \textit{et al.} \cite{ruan2019devil} also extends CE2P to M-CE2P to deal with the multiple human parsing problem. Specifically, two single human parsing models trained with detected and ground-truth human instances and one global human parsing model are fused to achieve the final instance-level human parsing results. Liu \textit{et al.} \cite{liu2019braidnet} proposes a braiding network which is consisted of two sub-networks, a semantic knowledge learning network and a local structure capturing network. Zhang \textit{et al.} \cite{zhang2020human} proposes a novel pyramidical gather-excite context network for the single human parsing task and extends it to multiple human parsing.
Ji \textit{et al.} \cite{ji2019learning} proposes to formulate the multiple human part regions into a semantic tree architecture, and a semantic aggregation module is also proposed to combine multiple hierarchical features and exploit the rich context information.
Li \textit{et al.} \cite{li2020self} proposes a model-agnostic self-correction mechanism for the human parsing task, which progressively refines labels in an online manner removing the influence coming from label noise.
The two-stage top-down methods are mainly devoted to training a robust model for single human parsing. Then human detection mainly relies on the extra Mask R-CNN \cite{he2017mask} detector.

\begin{figure*}[t]
  \centering
  \includegraphics[width=0.9\linewidth]{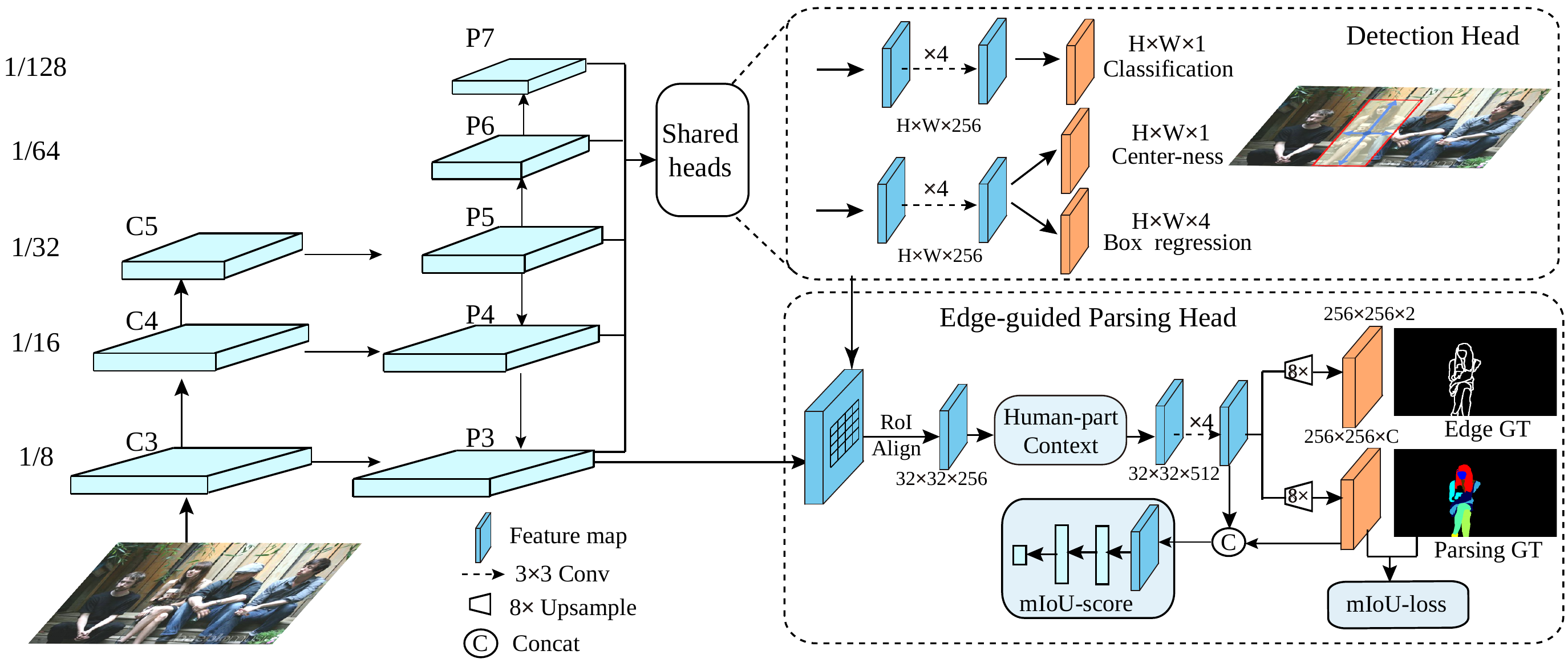}
  \caption{The overview of our proposed AIParsing framework. Given an input image, a feature pyramid network is firstly used along with a backbone CNN to extract multi-level feature maps. Then the multi-level features are fed to predict human instance bounding boxes through the detection head. And human instance parsing and edge maps are predicted via the edge-guided parsing head.}
  \label{fig:framework}
\end{figure*}

\noindent\textbf{One-stage top-down}: The difference between two-stage top-down and one-stage top-down lies in that whether the human instance detection and segmentation are designed in an end-to-end manner (as shown in Fig. \ref{fig:pipeline} (3)) or not. Although the two-stage top-down based models usually achieve powerful performance on instance-level human parsing,
the inference speed and flexibility is limited if it is deployed on practical systems. Thus, exploring one-stage top-down models is of practical significance.
Qin \textit{et al.} \cite{unified2019bmvc} proposes a top-down unified framework which simultaneously detects human instances and parse human parts within each instance with an attention module.
Zhao \textit{et al.} \cite{zhao2018understanding} proposes a top-down Nested Adversarial Network (NAN) for instance-level human parsing, which contains three GAN-like \cite{goodfellow2014generative} sub-nets, foreground prediction, instance-agnostic parsing, and instance-aware clustering. Yang \textit{et al.} \cite{yang2019parsing} proposes an one-stage top-down model named Parsing R-CNN for instance-level human analysis, which employs anchor-based approaches for human detection and a geometric and context encoding module for human instance parsing. Parsing R-CNN significantly improves the performance of the instance-level parsing metrics. Yang \textit{et al.} \cite{yang2020eccv} also extends Parsing R-CNN to a new RP R-CNN through introducing a global semantic network to enhance the parsing ability and a parsing re-scoring network to get high-quality parsing maps.
Our work follows the one-stage top-down design and explores a powerful anchor-free instance-level human parsing framework.

\section{Methodology}\label{method}

The overall pipeline of the Anchor-free Instance-level human Parsing network (\textbf{AIParsing}) is shown in Fig. \ref{fig:framework}. AIParsing contains three main parts: a Feature Pyramid Network (FPN), a human instance detection head, and an edge-guided human instance parsing head.
Given an input image, multi-level features are firstly extracted through the FPN-based backbone. Specifically, five levels of feature maps \{P3, P4, P5, P6, P7\} are used and they own five different strides 8, 16, 32, 64, and 128, respectively. P3, P4, and P5 are generated by the CNNs' feature maps C3, C4, and C5 followed by a $1 \times 1$ convolutional layer with the top-down connections \cite{lin2017feature}. P6 and P7 are produced via a convolutional layer with stride being 2 on P5 and P6, respectively. Then the multi-level features are used to predict human bounding boxes and parsing maps through the detection head and the edge-guided parsing head, respectively.

\subsection{Detection Head}
The goal of the detection head is to locate multiple human instances in the input image. In order to obtain accurate human instance bounding boxes, almost all existing instance-level human parsers, such as M-CE2P \cite{ruan2019devil}, SemaTree \cite{ji2019learning}, PGECNet \cite{zhang2020human}, Parsing R-CNN \cite{yang2019parsing}, choose anchor-based object detectors for human detection, \textit{i.e.}, Faster R-CNN \cite{ren2015faster}, which firstly generate a serial of proposals according to the pre-defined anchor boxes, then execute classification and offset regression. However, the pre-defined anchor boxes need tricky hyper-parameter tuning, which hampers the generality of new datasets or tasks.
Moreover, the lack of detail in predictions on a pixel level of two-stage anchor-based detectors does not harmonize with the following dense-pixel parsing task. At present, the Fully Convolutional Networks (FCNs) \cite{long2015fully} lead the mainstream frameworks in pixel-wise dense prediction tasks, such as semantic segmentation \cite{long2015fully} and scene parsing \cite{yuan2020object}.
For the instance-level human parsing task, the ultimate goal is to segment human instances with fine-grained categories.
Therefore, utilizing FCN-based architecture will be more effective since it enables incorporation of the instance detection and parsing tasks into a unified framework.
Motivated by FCOS \cite{tian2019fcos}, we introduce an anchor-free detector with fully convolutional operations as our human instance detector, which avoids parameter sensitivity related to anchor boxes. As shown in Fig. \ref{fig:framework}, the detection head is shared from P3 to P7, each position $(x,y)$ of the feature map $\text{P}_i$ predicts three outputs: a 4D box offset vector $\mathbf{t}^*$, a human classification score and a center-ness score.
The 4D bounding box regression vector $\mathbf{t}^*=(l^*, t^*, r^*, b^*)$ denotes the offset distances from the location ($x$, $y$) to the left, top, right, bottom sides of the bounding box, respectively. The location ($x$, $y$) is corresponding to the bounding box $B_i$ = $(x_{0}^{(i)}, y_{0}^{(i)}, x_{1}^{(i)}, y_{1}^{(i)}) \in \mathbb{R}^{4} $, where $(x_{0}^{(i)}, y_{0}^{(i)})$ and $(x_{1}^{(i)}, y_{1}^{(i)})$ denote the left-top and right-bottom points of the bounding box, and the 4D offset vector $\mathbf{t}^*$ is formulated as:

\begin{equation}
\begin{split}
l^{*} = x - x_{0}^{(i)}, t^{*} = y - y_{0}^{(i)}, \\
r^{*} = x_{1}^{(i)} - x, b^{*} = y_{1}^{(i)} - y.
\end{split}
\end{equation}

The loss of the detection head is:
\begin{equation}
L_{det} = L_{cls} + L_{reg} + L_{center},
\end{equation}
where $L_{cls}$ is the focal loss \cite{lin2017focal}, $L_{reg}$ is the bounding box IoU loss \cite{yu2016unitbox} and $L_{center}$ is the binary cross entropy loss. $L_{cls}$, $L_{reg}$ and $L_{center}$ are the loss of classification, offset regression and center-ness outputs, respectively.
The detection head extracts multiple human bounding boxes in the per-pixel prediction manner and feeds them into the following edge-guided parsing branch.

\begin{figure*}[t]
  \centering
  \includegraphics[width=0.9\linewidth]{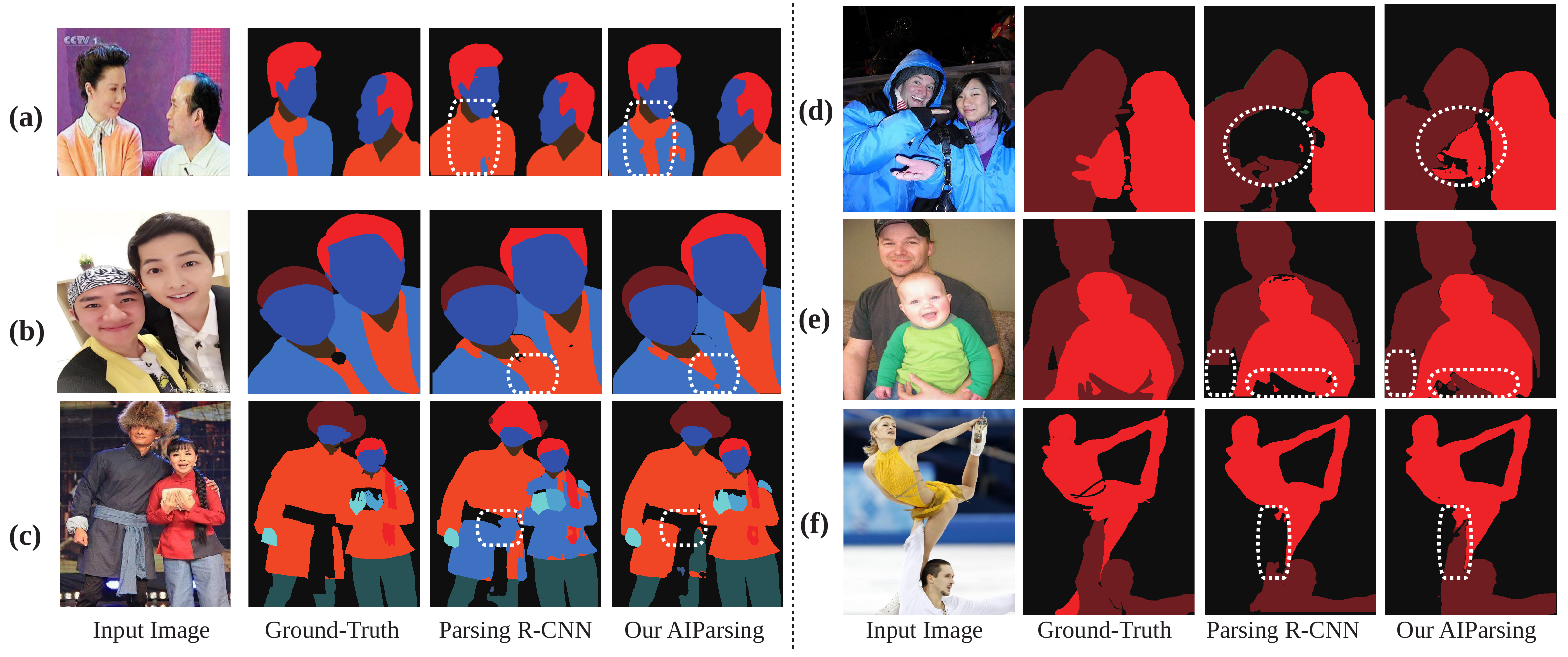}
  \caption{Illustration of the effectiveness of the edge clue. The first and second columns are the human instance masks and human parsing results, respectively. Both Parsing R-CNN \cite{yang2019parsing} and AIParsing use the same ResNet-50 backbone. Note that the AIParsing is employed without the refinement head for a fair comparison. }
  \label{fig:edge}

\end{figure*}

\subsection{Edge-guided Parsing Head}

The edge-guided human instance parsing head contains four main parts, \textit{i.e.}, detail-preserving, human-part context encoding, prediction head, and refinement head.

\noindent\textbf{Detail-preserving:}
As shown in Fig. \ref{fig:framework}, detail-preserving is performed on P3 via the RoIAlign operation, where P3 is the finest-resolution output feature map of the FPN. There are two reasons for adopting this mechanism according to the character of the instance-level human parsing task.
One reason is that we need enough annotated human instances with large sizes to train a robust model since small-size human instances have less appearance information. According to the statistics in the literature \cite{yang2019parsing}, $74\%$ and $86\%$ of human instances in CIHP \cite{gong2018instance} and LV-MHP v2.0 \cite{zhao2018understanding} datasets take up more than $10\%$ area of the image, respectively. The other reason is that
coarser-resolution feature maps provide limited appearance representation, especially for small human parts within the instance. If we perform RoI pooling on the coarser resolution (\textit{e.g.}, P7), some small human parts (for example, glove, left-hand, or glass) may be ignored in the coarse layers because of the down-sampling operations. Therefore, finer resolutions are more suitable for this task, as they can provide more detailed appearance information and better segmentation of human instances. Motivated by the high-resolution maintenance strategy used in DeepLabv3+ \cite{chen2018encoder},
which utilizes details from low-level feature maps (\textit{e.g.}, conv2) to recover detailed information, here we chose the finest-resolution P3 to perform RoI pooling. It seems insufficient that only performing RoI pooling on P3 may ignore the high-level semantic information. But in fact, the P3 feature map has fused the higher semantic information (C4 and C5) via the FPN, as shown in Fig. \ref{fig:framework}.

Similar to the feature map selection, the output size of RoI is also important.
Generally, the RoI pooling operation in Mask R-CNN \cite{he2017mask} projects an RoI region into a feature map with a fixed size of $7 \times 7$ (or $14 \times 14$). Since we need many details to densely parse the human instance, the desired output size of RoI should also preserve more instance details as possible. However, larger RoI resolution will lead to more computational cost, we chose $32 \times 32$ RoI output size as a trade-off in this paper.

\noindent\textbf{Human-part context encoding:}
The human-part context encoding module is implemented on the outputs of the detail-preserving step.
The context information is an important clue in semantic segmentation \cite{chen2017deeplab, qi2016hierarchically, chen2018encoder} and human parsing tasks \cite{ruan2019devil, zhang2020human, yang2019parsing, wang2020hierarchical, wang2021hierarchical}. There are two ways to explore the contextual information, \textit{i.e.}, the scale and multi-part relationship. The scale factor is focused on the multi-scale problem.
Human parts usually own various scales, hence an effective way to capture global and local information can help the following human parsing step. Indeed, the feature pyramid is a popular structure that can incorporate multiple scales.
Specifically, the Pyramidical Gather-Excite Context module (PGEC) \cite{zhang2020human} is adopted to detect multi-scale context.
Besides, the multi-part relationship provides valuable and relative information between various parts of the person.
The context of spatial position typically refers to a series of positions,
and the label of a spatial position (pixel) in an image denotes the category of the human part.
As a result, constructing a relationship between a spatial position and its context can reflect the relationship between different human parts.
Motivated by the self-attention mechanism which can capture long-range dependencies upon spatial positions, non-local operations \cite{wang2018non} are employed to explore the relationships between spatial positions.

At last, two types of context information are combined to formulate the human-part context encoding module, which provides rich context representation and helps to identify human parts. Specifically, PGEC contains one $3 \times 3$ convolution and four gather-excite units with spatial extent ratios $e = \{4, 8, 16, H\}$. A different point with the original PGEC is that we use five serial depth-wise convolution layers instead of the original one depth-wise convolution layer in GE\_H. The non-local operation chooses the embedded Gaussian version with group normalization \cite{wu2018group} in the last.

\noindent\textbf{Prediction head:}
The prediction head contains two tasks, parsing and edge predictions. One motivation for introducing extra edge prediction is to distinguish different human parts from each other in a single human instance. In general, there are various human parts within the human body region, and
accurately distinguishing adjacent human parts can assist with segmenting human parts well. As shown in the left column of Fig. \ref{fig:edge}, AIParsing can predict more accurate regions with the assistance of edge information than Parsing R-CNN. For example, in Fig. \ref{fig:edge} (a), two adjacent human parts (\textit{i.e.}, upper clothes and coat in the denoted white box) are hard to distinguish and while the Parsing R-CNN fails, AIParsing is able to predict accurate parsing results with the help of edge clue. Similar results can be observed from Fig. \ref{fig:edge} (b) and (c).
Another motivation is that there are various human instances in a detected bounding box, thus it is important to discriminate boundaries between overlapping instances. The edge information has been certificated in single human parsing \cite{ruan2019devil, zhang2020human} and semantic segmentation \cite{ding2019boundary, takikawa2019gated}, which is an useful clue to distinguish various human parts. The edge clue is also explored on instance segmentation tasks, for example, BMask R-CNN \cite{cheng2020boundary} learns object boundaries and masks in an end-to-end manner, which can predict
precise boundaries of masks. We extend it to help differentiate different human instances for locating accurate instance regions. Some examples can be referred to the right column of Fig. \ref{fig:edge}, our AIParsing predicts more precise instance regions compared to Parsing R-CNN. If two human instances share approximate appearances (Fig. \ref{fig:edge} (d)), or overlapping regions (Fig. \ref{fig:edge} (e)), or adjacent obscure boundaries (Fig. \ref{fig:edge} (f)), the proposed AIParsing can provide us with relatively complete regions.
The prediction head is implemented on the enhanced context representation. Specifically, a series of four shared convolution layers are conducted, and then two prediction branches, parsing and edge, are followed. The loss of the prediction head is:
\begin{equation}
L_{pred} = \alpha L_{parsing} + \beta L_{edge},
\end{equation}
where $\alpha=\beta=2$. $L_{parsing}$ is the standard cross-entropy loss. $L_{edge}$ is the weighted cross-entropy loss, the formulation is:

\begin{equation}
L_{edge} = - w_{0} \sum_{i \in Y_{-}} log (p_{i}(y_{i} = 0)) - w_{1} \sum_{i \in Y_{+}} log (p_{i}(y_{i} = 1)),
\end{equation}
where $w_{0} = \frac{|Y_{+}|}{|Y|}$, $w_{1} = \frac{|Y_{-}|}{|Y|}$. $Y_{-}$ and $Y_{+}$ are the ground-truth pixels belonging to non-edge and edge, respectively. $p_{i}(\cdot)$ denotes the probability of $i$-th pixel.

\begin{table*}
\centering
\caption{Performance comparison in terms of mIoU, $\text{AP}_{\text{vol}}^{\text{p}}$, $\text{AP}_{\text{50}}^{\text{p}}$ and $\text{PCP}_{\text{50}}$ with state-of-the-art methods on the validation set of CIHP \cite{gong2018instance}. * denotes using COCO pre-training \cite{lin2014microsoft}. $\dag$ denotes test-time augmentation.}
 \label{table:CIHPall}
\small{

\begin{tabular}{l | c | c | c c c c }
\hline
  Method &  Backbone  & Epoch & mIoU & $\text{AP}_{\text{vol}}^{\text{p}}$ & $\text{AP}_{\text{50}}^{\text{p}}$ & $\text{PCP}_{\text{50}}$ \\
 \hline

 Bottom-up:  & & & & & &  \\

  \quad PGN$\dag$ \cite{gong2018instance}  & ResNet-101 & 80 & 57.3 & 35.1 & 37.5 & - \\
  \quad DeepLab v3+$\dag$  \cite{chen2018encoder} & Xception & 100& 58.9 & - & - & - \\
  \quad Graphonomy$\dag$ \cite{gong2019graphonomy} & Xception  & 100& 58.6 & - & - & - \\
  \quad CorrPM \cite{zhang2020CVPR} & ResNet-101 & 150& 60.2 & - & - & - \\
  \quad GPM$\dag$  \cite{he2019grapy} & Xception  & 100& 60.3 & - & - & - \\
  \quad Gray\_ML$\dag$  \cite{he2019grapy} & Xception & 200& 60.6 & - & - & - \\
  \quad PCNet \cite{zhang2020part} & ResNet-101 & 120 & 61.1 & - & - & - \\
\hline
Two-stage top-down:  & & & & & & \\
 \quad BraidNet \cite{ruan2019devil} & ResNet-101 &150& 60.6 & - & - & - \\
 \quad SemaTree \cite{ji2019learning} & ResNet-101 & 200& 60.9 & - & - & - \\
 \quad M-CE2P \cite{ruan2019devil} & ResNet-101 & 150&63.6 & 48.9 & 54.7 & - \\
 \quad PGECNet \cite{zhang2020human} & ResNet-101  & 150&64.6& - & - & - \\
 \quad SCHP$\dag$ \cite{li2020self} & ResNet-101  & 150&\textbf{67.5} & 52.0 & 58.9 & - \\
\hline
One-stage top-down:  & & & & & & \\
 \quad Unified \cite{unified2019bmvc} & ResNet-101  & 37 & 55.2 & 48.0 & 51.0 & - \\
 \quad Parsing R-CNN \cite{yang2019parsing} & ResNet-50  & 75 &56.3 & 53.9 & 63.7 & 60.1 \\
 \quad  Mask Scoring \cite{huang2019mask} & ResNet-50& 75 &  56.0 & 56.8  & 68.9   & 59.9 \\
 \quad BMask R-CNN \cite{cheng2020boundary} & ResNet-50 & 75 & 56.4 & 54.3  & 64.6   & 61.8 \\

 \quad Parsing R-CNN* \cite{yang2019parsing} & ResNet-50  & 75 &57.5 & 54.6 & 65.4  & 62.6 \\
 \quad RP RCNN \cite{yang2020eccv} & ResNet-50  &75 & 58.2&  58.3 & 71.6 & 62.2 \\

 \quad \textbf{AIParsing} & ResNet-50  & 75 & 59.7 & 59.2 & 73.1 & 66.3 \\

 \quad Parsing R-CNN* \cite{yang2019parsing} & ResNeXt-101  & 75 &59.8 & 55.9 & 69.1 & 66.2 \\
\quad \textbf{AIParsing} & ResNet-101 & 75 & 60.7 & 60.3  & 75.2& 68.5 \\
 \quad \textbf{AIParsing} & ResNeXt-101 & 75 & 60.8 & 60.5 & 76.0 & 69.2 \\
  \quad Parsing R-CNN*$\dag$ \cite{yang2019parsing} & ResNeXt-101  &75 & 61.1 & 56.5 & 71.2 & 67.7 \\
   \quad RP RCNN$\dag$ \cite{yang2020eccv} & ResNet-50  & 150 &61.8 & 61.2 & 77.2 & 70.5 \\
  \quad \textbf{AIParsing}$\dag$ & ResNet-50  & 75 & 62.5 & 62.2 & 78.3 & 71.6 \\
 \quad \textbf{AIParsing}$\dag$ & ResNet-101 & 75  & 63.0 &  63.1 & 79.2 & 71.7 \\
 \quad \textbf{AIParsing}$\dag$ & ResNeXt-101  & 75 &63.3 & \textbf{63.4} & \textbf{79.9} & \textbf{72.3} \\
\hline
\end{tabular}

}
\end{table*}

\noindent\textbf{Refinement head:}
While the parsing prediction results are obtained through the prediction head, they are still not ideal. There are two main problems upon the predicted bounding boxes, \textit{i.e.}, low-quality global-level box and imprecise part-level parsing map. The low-quality predicted bounding boxes with inaccurate regions will lead to lower IoU scores, and influence the instance-level AP scores. Meanwhile, if the parsing maps are with imprecise predictions, they will cause lower IoU scores on some human classes, and will result in an unsatisfactory global-level parsing mIoU score. Thus, these two issues should be fused in a complementary manner, where the global-level box focuses on improving the quality of the predicted bounding boxes, and the part-level parsing maps focus on improving the quality of each human class. Specifically, the global-level boxes are optimized through a re-scoring subnetwork estimating the mIoU score of the predicted human instance parsing map within the detected bounding box, where the subnetwork consists of two convolution layers and three fully connected layers.
The part-level parsing results are implemented via optimizing a structure-aware tractable surrogate loss (mIoU loss) \cite{berman2018lovasz}.
The loss of the refinement head is:

\begin{equation}
L_{refine} = \theta L_{miou} + \gamma L_{miou-score},
\end{equation}
where $L_{miou}$ is the parsing mIoU loss \cite{berman2018lovasz}, and $L_{miou-score}$ adopts the MSE loss, $\gamma=1, \theta=2$.

The total loss of AIParsing $L_{total}$ is the sum of the detection head $L_{det}$, the prediction head $L_{pred}$, and the refinement head $L_{refine}$.

\section{Experiments}\label{experiment}

In this section, we evaluate the proposed Anchor-free Instance-level human Parsing network (AIParsing) on two popular image instance-level human parsing datasets, \textit{i.e.}, CIHP \cite{gong2018instance} and LV-MHP-v2.0 \cite{zhao2018understanding}, and one video multiple human parsing benchmark, \textit{i.e.}, VIP \cite{zhou2018adaptive}.

\subsection{Datasets and Metrics}

\noindent\textbf{CIHP}: Crowd Instance-level Human Parsing (CIHP) \cite{gong2018instance} is a large-scale multiple human parsing dataset in the wild. There are $38,280$ human images in total, which are annotated with $19$ pixel-wise human semantic parts and instance-level human IDs.
The whole dataset is split into three parts: $28,280$ images for training, $5,000$ images for validation, and $5,000$ images for testing, respectively.

\noindent\textbf{LV-MHP-v2.0}: Learning Vision Multi-Human Parsing (LV-MHP-v2.0) \cite{zhao2018understanding} is also a large-scale challenge instance-level human parsing dataset captured in real-world scenes from various conditions.
LV-MHP-v2.0 contains 25,403 images, each image is annotated with 58 fine-grained semantic category labels and contains 2-26 persons. 15,403, 5000, 5000 images are used for training, validation, and testing, respectively.

\noindent\textbf{VIP}: Video Instance-level Parsing (VIP) \cite{zhou2018adaptive} dataset is a large-scale video-based multiple human parsing benchmark. There are 404 videos in total and 20 annotated semantic part labels (same with CIHP). VIP is divided into three parts, 304 videos for training, 50 videos for validation, and 50 videos for testing, respectively. The frames in the training and validation set are $16,024$, and $2,445$, respectively.

\noindent\textbf{Metrics}: Two types of metrics are evaluated, \textit{i.e.}, global-level and instance-level metrics. For the global parsing metric, the standard mean Intersection over Union (mIoU) score is reported. For local-sensitive instance-level evaluation, the Average Precision based on part ($\text{AP}^{\text{p}}$) is adopted to evaluate the instance-level human parsing performance. Specifically, $\text{AP}_{\text{vol}}^{\text{p}}$ and $\text{AP}_{\text{50}}^{\text{p}}$ scores are reported, where $\text{AP}_{\text{50}}^{\text{p}}$ denotes the IoU threshold equal to 0.5 and $\text{AP}_{\text{vol}}^{\text{p}}$ is the mean of various IoU thresholds (IoU $\in$ [0.1, 0.9], with 0.1 increment).
In addition, the Percentage of Correctly parsed semantic Parts (PCP) metric is used to evaluate the quality within the human instance. The PCP of one human instance is the ratio between the correctly parsed human semantic part number and the total human semantic part number of that human. Specifically, $\text{PCP}_{\text{50}}$ score is reported. While for video human parsing task, we also adopt another instance-level metric, \textit{i.e.}, average precision based on region ($\text{AP}^{\text{r}}$), which is used to evaluate the performance from the view of each class.

\noindent\textbf{Implementation details}: We follow almost all the hyper-parameters in FCOS \cite{tian2019fcos} except for replacing positive score threshold value 0.03 with 0.05 as suggested in \cite{lee2020centermask}. FPN levels P3 to P7 are with 256 channels. RoIAlign is selected as the RoI pooling operation.
In the training stage, the input image scales are randomly sampled from [640, 800] pixels. Then the images are resized to 800 pixels on the short side and 1,333 pixels or less on the long side \cite{tian2019fcos, lee2020centermask}. AIParsing is trained with Stochastic Gradient Descent (SGD) for about $75$ epochs with a mini-batch of 8 images and an initial learning rate of 0.005 then is decreased by a factor of 10 at 50 and 65 epochs, respectively. The weight decay and momentum are set as 0.0001 and 0.9, respectively.
We initialize the backbone models with ImageNet pre-trained weights \cite{deng2009imagenet}. In the inference stage, the human detection head with FCOS generates 50 high-score human instance boxes, and these boxes are fed into the edge-guided parsing head to predict parsing maps on each RoI.
When we inference with test-time augmentation, the shorter side is resized at multiple scales and keep the aspect ratio, and the horizontal flipping is also applied. For CIHP and VIP, we use multiple scales of (500, 600, 700, 800, 850). For LV-MHP-v2.0, the scales are (500, 600, 700, 800, 850, 1000).

\begin{table*}
\centering
\caption{ Performance comparison in terms of mIoU, $\text{AP}_{\text{vol}}^{\text{p}}$, $\text{AP}_{\text{50}}^{\text{p}}$ and $\text{PCP}_{\text{50}}$ with state-of-the-art methods on the validation set of LV-MHP-v2.0 \cite{zhao2018understanding}. * denotes using COCO pre-training \cite{lin2014microsoft}. $\dag$ denotes test-time augmentation.}
 \label{table:LVall}
\small{

\begin{tabular}{l | c | c | c c c c }
\hline
  Method &  Backbone  & Epoch & mIoU & $\text{AP}_{\text{vol}}^{\text{p}}$ & $\text{AP}_{\text{50}}^{\text{p}}$ & $\text{PCP}_{\text{50}}$ \\
 \hline
 Bottom-up:  & & & & &  &  \\
 \quad MH-Parser \cite{li2017multiple}  & ResNet-101 & - & - & 36.1 & 18.0 & 27.0 \\
 \quad NAN \cite{zhao2018understanding}  & - & 80 &  - & 41.7 & 25.1 & 32.2 \\
 \quad MG-Parsing \cite{zhou2021multi} & ResNet-101 & 150 & 41.4 & 44.3 & 39.0 & 42.3 \\
\hline
Two-stage top-down:  & & & & & &  \\
\quad SemaTree \cite{ji2019learning} & ResNet-101 & 200 & - & 42.5 & 34.4 & 43.5 \\
\quad M-CE2P \cite{ruan2019devil} & ResNet-101  & 150 &41.1  & 42.7& 34.5 & 43.8 \\
\quad SCHP$\dag$ \cite{li2020self} & ResNet-101  & 150&\textbf{45.2} & 45.3 & 35.1 & 48.0 \\
\hline
One-stage top-down:  & & & & & &   \\
\quad Mask R-CNN \cite{he2017mask} & ResNet-50 & - & - & 33.8 & 14.9 & 25.1 \\
\quad Parsing R-CNN \cite{yang2019parsing} & ResNet-50 & 75 & 36.2 & 39.5 & 24.5 & 37.2 \\
\quad Parsing R-CNN* \cite{yang2019parsing} & ResNet-50 & 75 & 37.0 & 40.3 & 26.6  & 40.0 \\

 \quad RP RCNN \cite{yang2020eccv} & ResNet-50  &75 & 37.3 &  45.2 & 40.5 & 39.2 \\
\quad \textbf{AIParsing} & ResNet-50 & 75 & 39.9 & 45.9 & 41.1 & 45.3 \\
\quad Parsing R-CNN* \cite{yang2019parsing} & ResNeXt-101 & 75 & 40.3 & 41.8 & 30.2 & 44.2 \\
\quad Parsing R-CNN*$\dag$ \cite{yang2019parsing} & ResNeXt-101 & 75 & 41.8 & 42.7 & 32.5 & 47.9 \\
\quad \textbf{AIParsing} & ResNet-101 & 75 & 40.1 & 46.6  & 43.2& 47.3 \\
\quad \textbf{AIParsing} & ResNeXt-101 & 75 & 40.4 & 47.5  & 45.0& 48.7 \\

 \quad \textbf{AIParsing}$\dag$ & ResNet-101 & 75 & 42.1 &  49.3 & 51.3 & 54.9 \\

\quad \textbf{AIParsing}$\dag$ & ResNeXt-101 & 75 & 42.4 & \textbf{49.9} & \textbf{52.8} & \textbf{56.1} \\
\hline
\end{tabular}

}
\end{table*}

\begin{table*}
\centering
\caption{ Performance comparison with state-of-the-art methods on the validation set of VIP \cite{zhou2018adaptive}. * denotes directly tested on the well-trained AIParsing model with CIHP dataset.}
 \label{table:VIPall}
\small{

\begin{tabular}{l | c | c c c c c c | c c c }
\hline
  Method &  mIoU  & $\text{AP}_{\text{50}}^{\text{r}}$ & $\text{AP}_{\text{60}}^{\text{r}}$ & $\text{AP}_{\text{70}}^{\text{r}}$  & $\text{AP}_{\text{80}}^{\text{r}}$
  & $\text{AP}_{\text{90}}^{\text{r}}$ & $\text{AP}_{\text{vol}}^{\text{r}}$
   &$\text{AP}_{\text{vol}}^{\text{p}}$ & $\text{AP}_{\text{50}}^{\text{p}}$ & $\text{PCP}_{\text{50}}$ \\
 \hline

 \quad DFF \cite{zhu2017deep}  & 35.3 & 20.3 & 15.0 & 9.8 & - & -& 20.3 & - & -& -\\
 \quad FGFA \cite{zhu2017flow}  & 37.5 &  24.0 & 17.8 & 12.2 & - & -& 23.0 & - & - &- \\
 \quad ATEN \cite{zhou2018adaptive} & 37.9 & 25.1 & 18.9 & 12.8& - & -& 24.1 & - & - & -\\
 \quad SCHP \cite{li2020self} & \textbf{63.2} & 57.8 & 52.9 & 46.3 & 34.1 & 12.4 & 51.4 & - & - & - \\
\hline

\quad \textbf{AIParsing*} & 60.4& 58.9 & 53.9 & 45.9 & 31.6 & 10.5 & 51.5 & 63.8 & 80.7 & 70.2 \\

\quad \textbf{AIParsing} & 61.4 & \textbf{59.3} & \textbf{54.2} & \textbf{47.7}  & \textbf{34.5} & \textbf{12.6} & \textbf{52.2} & \textbf{64.2} & \textbf{81.6} & \textbf{71.1} \\

\hline
\end{tabular}

}
\end{table*}

\subsection{Comparison with State-of-the-art Methods}

To validate the effectiveness of AIParsing, we analyzed the comparison results with state-of-the-art bottom-up, one-stage, and two-stage top-down parsing methods on CIHP \cite{gong2018instance} and LV-MHP-v2.0 \cite{zhao2018understanding} validation sets, respectively. In addition, we also certificate our proposed AIParsing on the video human parsing dataset, \textit{i.e.}, VIP \cite{zhou2018adaptive}.

\noindent\textbf{CIHP}: We analyzed the comparison results from two views, the global-level metric (mIoU) and the instance-level metrics ($\text{AP}_{\text{vol}}^{\text{p}}$, $\text{AP}_{\text{50}}^{\text{p}}$ and $\text{PCP}_{\text{50}}$).

\textbf{1) For the global parsing metric}, as shown in Table \ref{table:CIHPall}, the proposed AIParsing outperforms all one-stage top-down methods in terms of mIoU score.
While using the same backbone ResNet-50, the proposed AIParsing achieves 1.5\% performance improvement compared to RP R-CNN \cite{yang2020eccv} in terms of mIoU score. Parsing R-CNN adopts MS-COCO \cite{lin2014microsoft} pre-training to enhance the parsing ability, conversely, our proposed AIParsing without COCO pre-training can obtain better performance with ResNeXt-101-32x8d backbone (60.8\% vs. 59.8\%) comparing with Parsing R-CNN. While using test-time augmentation, AIParsing achieves the best mIoU score of 63.3\% over all one-stage top-down ones. We can find that the mIoU score of AIParsing is inferior to the best two-stage top-down method SCHP \cite{li2020self}. The best mIoU performance shows that the two-stage top-down methods are good at obtaining global parsing results. It benefits from the two-stage top-down method that improves the global parsing performance through
training three independent parsing models, \textit{i.e.}, one global parsing model and two instance-level parsing models. The final global parsing results are acquired by fusing these three parsing models in the inference step. However, two-stage top-down methods are not end-to-end, which limits the practicability in the real-world scene. Bottom-up and one-stage top-down methods are end-to-end, which are encouraged for practical applications. The state-of-the-art one-stage top-down method RP R-CNN \cite{yang2020eccv} and our AIParsing both attain better mIoU scores than all bottom-up methods.
Consequently, the proposed AIParsing is effective in obtaining better global parsing performance, it also owns higher instance-level metric scores. However, the two-stage top-down methods fall drastically on instance-level metrics, we will analyze the reasons in the following.

\textbf{2) For the instance-level parsing metrics}, as shown in Table \ref{table:CIHPall}, RP R-CNN \cite{yang2020eccv} with ResNet-50 backbone achieves 6.3\% and 12.7\% gains than SCHP \cite{li2020self} with ResNet-101 backbone in terms of $\text{AP}_{\text{vol}}^{\text{p}}$ and $\text{AP}_{\text{50}}^{\text{p}}$, respectively. The reason lies in that human detection and parsing steps of two-stage top-down methods are separated, and the detector depends on extra well-trained object detectors without further training. Thus, these two steps trained in parallel in one-stage top-down methods will obtain better instance-level metric scores.
With the same ResNet-50 backbone, AIParsing obtains state-of-the-art results and outperforms the runner-up RP R-CNN \cite{yang2020eccv} by large margins of 0.9\% $\text{AP}_{\text{vol}}^{\text{p}}$ / 1.5\% $\text{AP}_{\text{50}}^{\text{p}}$ / 4.1\% $\text{PCP}_{\text{50}}$. These large margins benefit from the effectiveness of the anchor-free motivation and the useful design of the edge-guided human instance parsing branch. The large superiority still exists on test-time augmentation results.
While using a deeper ResNeXt-101-32x8d backbone, AIParsing attains total suppression compared to Parsing R-CNN in terms of all instance-level parsing metrics. The proposed AIParsing with single-scale testing even outperforms Parsing R-CNN with multi-scale testing
both on $\text{AP}_{\text{vol}}^{\text{p}}$ and $\text{PCP}_{\text{50}}$. Further, AIParsing with multi-scale testing obtains the best instance-level parsing metric scores upon all state-of-the-art methods and the satisfactory global-level mIoU score.

\noindent\textbf{LV-MHP-v2.0}: We also report the evaluation metric scores of the AIParsing on the LV-MHP-v2.0 validation set, as shown in Table \ref{table:LVall}.
The proposed AIParsing equipped with ResNet-50 backbone can achieve better performance than RP R-CNN \cite{yang2020eccv} and Parsing R-CNN \cite{yang2019parsing} on both global-level and instance-level metrics. Specifically, AIParsing outperforms 2.6\% mIoU than RP R-CNN \cite{yang2020eccv} with the same ResNet-50 backbone.
When a deeper ResNeXt-101-32x8d backbone with test-time augmentation is adopted, AIParsing also obtains the best performance with 42.4\% / mIoU, 49.9\% / $\text{AP}_{\text{vol}}^{\text{p}}$, and 56.1\% / $\text{PCP}_{\text{50}}$.

\noindent\textbf{VIP}: Except image instance-level human parsing data, we further report the comparison metric scores of AIParsing on the video human parsing data, \textit{i.e.}, VIP validation set, as shown in Table \ref{table:VIPall}. Because CIHP and VIP share the same human parts, thus we test the performance on VIP val directly adopting the pre-trained AIParsing model with CIHP dataset (denoted it as AIParsing*). The performance ($\text{AP}_{\text{vol}}^{\text{r}}$) of AIParsing* is close to current state-of-the-art method SCHP, which shows the generalization ability of the AIParsing. Further, we also provide the results of AIParsing trained with VIP training set. AIParsing can obtain the best instance-level metric score, \textit{i.e.}, 52.2\% $\text{AP}_{\text{vol}}^{\text{r}}$. The observations from VIP are similar to the image datasets where the proposed AIParsing owns superiority on the instance-level metrics.

\begin{table}
\centering
\caption{ Anchor-based vs. Anchor-free detectors comparison on the validation set of CIHP \cite{gong2018instance}.}
 \label{table:anchor}
\small{
\begin{tabular}{c| c | c c c c }
\hline
  Method & $\text{mAP}^{\text{bbox}}$ & mIoU & $\text{AP}_{\text{vol}}^{\text{p}}$  & $\text{PCP}_{\text{50}}$ \\
 \hline
 RPN-parsing  & 65.8 & 52.8 & 51.2 & 55.3 \\
 Anchor-free-parsing & \textbf{71.4} & \textbf{54.0} &  \textbf{52.5}  & \textbf{59.8} \\

\hline
\end{tabular}

}
\end{table}

\begin{table}
\centering
 \caption{Different RoI output sizes comparison on the validation set of CIHP \cite{gong2018instance}. FLOPs is only computed according to the multi-scale context module PGEC.}
 \label{table:roi}
\small{
\begin{tabular}{c| c | c | c c c c }
\hline
  RoI size & FLOPs(G) & mIoU & $\text{AP}_{\text{vol}}^{\text{p}}$ & $\text{AP}_{\text{50}}^{\text{p}}$ & $\text{PCP}_{\text{50}}$ \\
 \hline
 $14 \times 14$  & \textbf{1.2}  &  51.4 & 50.0  & 54.4   &  55.0 \\
 $32 \times 32$ & 6.4  & 54.0 &  52.6 & 60.5 & 59.6 \\
$48 \times 48$ & 14.5  & \textbf{54.7}  &  \textbf{53.6} & \textbf{62.9}  & \textbf{61.1}  \\

\hline
\end{tabular}

}

\end{table}

\begin{table*}[tb]
\centering
\caption{ Different components comparison of the edge-guided parsing branch on the validation set of CIHP \cite{gong2018instance}.
}
 \label{table:edge}
\small{
\begin{tabular}{c |c c| c c | c c | c | c c c c }
 \hline
   \multirow{2}*{ResNet50-FPN } & \multicolumn{2}{c|}{Human-part context }& \multirow{2}*{Edge} &\multirow{2}*{GN} &\multicolumn{2}{c|}{Refinement head } & \multirow{2}*{3x} & \multirow{2}*{mIoU}&\multirow{2}*{$\text{AP}_{\text{vol}}^{\text{p}}$} & \multirow{2}*{$\text{AP}_{\text{50}}^{\text{p}}$ }& \multirow{2}*{$\text{PCP}_{\text{50}}$}\\
 \cline{2-3} \cline{6-7}
    & PGEC & Non-local & & &  mIoU-loss & mIoU-score & & & & & \\


 \hline
 $\checkmark$ &  & & & & & & & 49.6 & 45.8  & 43.1 & 47.9 \\
 $\checkmark$ & $\checkmark$ & & & & & & & 52.5 & 50.8  & 55.6 & 56.6 \\
 $\checkmark$ &  & $\checkmark$ & & & & & & 52.7 & 50.7  & 55.7 & 56.0 \\
 $\checkmark$ & $\checkmark$ &$\checkmark$  & & & & & & 54.0 &  52.6 & 60.5 & 59.6 \\
$\checkmark$ & $\checkmark$ & $\checkmark$  & $\checkmark$ & & & & & 54.8 &  53.8 & 62.7 & 61.5 \\
$\checkmark$ & $\checkmark$ & $\checkmark$  & $\checkmark$& $\checkmark$& & & & 55.6 &  54.4 & 64.4 & 62.8 \\
$\checkmark$ & $\checkmark$ & $\checkmark$  & $\checkmark$& $\checkmark$& $\checkmark$& & & 57.6 & 54.4 & 64.8 & 63.9 \\
$\checkmark$ & $\checkmark$ & $\checkmark$  & $\checkmark$& $\checkmark$& & $\checkmark$ & & 55.8 & 56.8 & 69.1 & 62.1 \\
$\checkmark$ & $\checkmark$ & $\checkmark$  & $\checkmark$& $\checkmark$& $\checkmark$ & $\checkmark$ & & 57.3 & 56.9 & 69.4 & 64.4 \\
$\checkmark$ & $\checkmark$ & $\checkmark$  & $\checkmark$& $\checkmark$& $\checkmark$&$\checkmark$ &$\checkmark$ & \textbf{59.7} & \textbf{59.2} & \textbf{73.1} & \textbf{66.3} \\

\hline
\end{tabular}

}
\end{table*}

\subsection{Ablation Study}
In this section, we analyze each part's effectiveness of AIParsing.
Unless specified, all ablation studies are conducted on the CIHP validation set using AIParsing with ResNet-50-FPN backbone and trained for 25 epochs.

\noindent\textbf{Anchor-based detector vs. Anchor-free detector:}
We use the anchor-based human parsing model Parsing R-CNN \cite{yang2019parsing} as our baseline in this setting, and we directly borrow the evaluation score from literature \cite{yang2019parsing}, which is trained for 25 epochs, and we denote it as RPN-parsing.
For further comparison, we just replaced the RPN in Parsing R-CNN with an anchor-free FCOS detector while other parts are kept the same, and denote it as Anchor-free-parsing. As shown in Table \ref{table:anchor}, anchor-free-parsing model improves the $\text{mAP}^{\text{bbox}}$ score by about 5.6\% compared with the RPN-parsing model, thus shows that the one-stage anchor-free human detector FCOS can achieve better performance than the anchor-based detector. When the detection model obtains better performance, this will help to improve the performance of global-level and instance-level parsing metrics, and thus Anchor-free-parsing obtains higher $\text{AP}_{\text{vol}}^{\text{p}}$ and $\text{PCP}_{\text{50}}$ scores than RPN-parsing. Besides, its per-pixel design style is consistent with the following human instance parsing task.

\noindent\textbf{RoI output size:}
The RoI size is important for dense human instance parsing since parsing needs sufficient human details to distinguish various parts of a human instance. From our common sense, a larger RoI size owns more details than a small one. However, a large RoI size will generate more memory and computational cost. In Table \ref{table:roi}, three different RoI sizes, \textit{i.e.}, $14 \times 14$, $32 \times 32$, and $48 \times 48$, are employed to validate. Although the $14 \times 14$ size holds fewer computations, it obtains lower performance than the $32 \times 32$ on all metrics. With a larger $48 \times 48$, it can obtain better scores than the $32 \times 32$. However, the large size needs more computational cost by observing the FLOPs. Therefore, we choose the RoI size $32 \times 32$ as a trade-off.

\noindent\textbf{Human-part context:}
Human-part context is an effective module in capturing multi-scale context information (PGEC) and the relationship of various human parts (non-local). In Table \ref{table:edge}, both the PGEC part and the non-local part have the ability to on improving all metric scores. This shows the effectiveness of each part in the human-part context module. If these two parts are simultaneously added, a 4.4\% improvement in terms of mIoU score is achieved. For instance-level metrics, the human-part context module (PGEC+Non-local) obtains 6.8\%, 17.4\%, 11.7\% improvements than the baseline model in terms of $\text{AP}_{\text{vol}}^{\text{p}}$, $\text{AP}_{\text{50}}^{\text{p}}$ and $\text{PCP}_{\text{50}}$, respectively. It shows the effectiveness of exploiting scale and relationship clues to assist in human instance parsing.

\noindent\textbf{Edge information:}
Introducing edge clues is useful to distinguish adjacent human semantic parts and different instances. The edge label can be obtained by pre-processing the labeled human instance parsing map without extra annotation cost. Specifically, we scan and apply the XOR operation to the neighboring pixels, and use a $3 \times 3$ kernel implemented on the parsing label to extract the edge pixels. From the performance comparison results in Table \ref{table:edge}, the edge information can achieve 0.8\% improvement than human-part context based method in terms of mIoU. Meanwhile, instance-level parsing metrics also have 1.2\%, 2.2\%, 1.9\% improvements than human-part context based method in terms of $\text{AP}_{\text{vol}}^{\text{p}}$, $\text{AP}_{\text{50}}^{\text{p}}$, and $\text{PCP}_{\text{50}}$, respectively.

\noindent\textbf{Group normalization:}
To obtain better performance, group normalization (GN) \cite{wu2018group} is introduced to ensure the stability of the training stage. We add the GN operation after the last convolutional layer of the prediction head. Specifically, the method with GN improves 0.8\%, 0.6\%, 1.7\% and 1.3\% compared with the ablative method without GN in terms of mIoU, $\text{AP}_{\text{vol}}^{\text{p}}$, $\text{AP}_{\text{50}}^{\text{p}}$, and $\text{PCP}_{\text{50}}$, respectively. The good performance shows the effectiveness of introducing group normalization into the parsing module.

\begin{figure*}[t]
  \centering
  \includegraphics[width=0.85\linewidth]{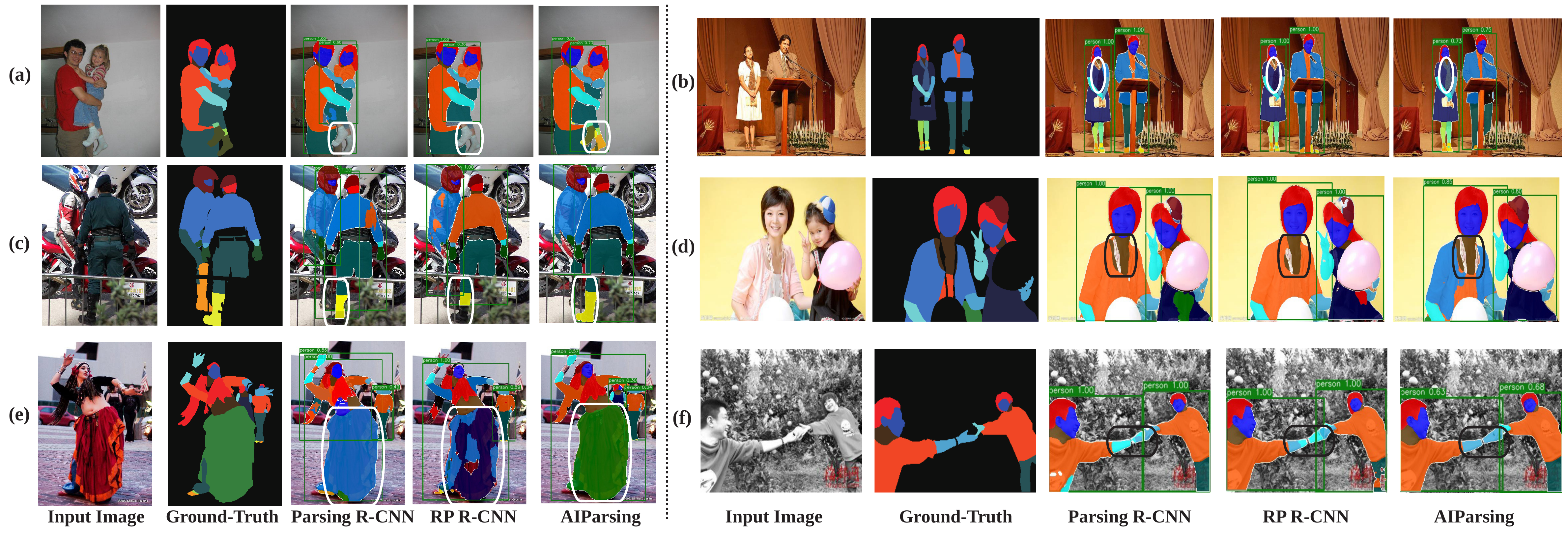}
  \caption{Visual comparison on CIHP val. AIParsing obtains accurate predictions than Parsing R-CNN \cite{yang2019parsing} and RP R-CNN \cite{yang2020eccv} equipped with same ResNet-50 backbone. The regions labeled in red boxes denote the improved parsing results by our AIParsing.
   }
  \label{fig:three}
\end{figure*}

\noindent\textbf{Refinement head:} As show in Table \ref{table:edge}, if we employ box-level mIoU score to filter out low-quality bounding boxes, it will give more helps on instance-level metric scores, \textit{i.e.}, $\text{AP}_{\text{vol}}^{\text{p}}$ and $\text{AP}_{\text{50}}^{\text{p}}$. The reason lies in that some detected bounding boxes own high confidence scores but low-quality predicting results, these boxes will hurt the mIoU score and further lead to unideal $\text{AP}^{\text{p}}$ score. While we combine box-level mIoU score and confidence score to estimate the quality of the detected bounding boxes, these poorly predicted bounding boxes will be discarded. Thus, the $\text{AP}^{\text{p}}$ score obtains improvement. However, the optimization of mIoU-score only considers the global-level quality of the detected bounding boxes, but the parsing quality of each part within the detected ones is also important.
When we add the part-level constraint, \textit{i.e.}, equipped with mIoU-loss, the method improves 2.0\% mIoU score compared with without this constraint. The advantage of the mIoU metric score proves that the high-quality parsing map assists in obtaining better global parsing results. Here, the $\text{AP}^{\text{p}}$ scores are close whether with mIoU-loss constraint or not, the main reason lies in that the selected low-quality detected bounding boxes with high confidence scores lead to approximate scores.
When we combine these two useful parts together, it obtains 1.7\%, 2.5\%, 5.0\%, 1.6\% improvements than the ablative method without them in terms of mIoU, $\text{AP}_{\text{vol}}^{\text{p}}$, $\text{AP}_{\text{50}}^{\text{p}}$, and $\text{PCP}_{\text{50}}$, respectively. The good performance of the refinement head encourages us to taking these clues into account.

\noindent\textbf{Longer schedule:}
Increasing iterations or epochs is a useful mechanism to improve performance \cite{yang2019parsing}. As shown in Table \ref{table:edge}, we report the performance of adopting three times standard training schedule (25 epochs, denoted as $1\times$ ), \textit{i.e.}, 75 epochs in total, denoted as $3\times$. While training the model with $3\times$ epochs, the mIoU performance is improved by 2.4\% compared with that of $1\times$ epoch.

\noindent\textbf{Effectiveness of GN and 3x:}
For a fair comparison, we chose Parsing R-CNN \cite{yang2019parsing} trained with $3\times$ (75 epochs) and GN as the baselines, while the proposed AIParsing method discards the refinement head, as shown in Table \ref{table:gn3x}. If we replace the Parsing R-CNN with the same FCOS detector, the detection score $\text{mAP}^{\text{bbox}}$, AP and PCP scores can be boosted much more. Moreover, if we add the edge-guided parsing head without containing the refinement part, AIParsing(w/o refine) outperforms Parsing R-CNN in all metric scores.

\begin{table}

\centering
\footnotesize{
\caption{Performance comparison on CIHP val. * denotes the results from \cite{yang2020eccv}.}
 \label{table:gn3x}
\begin{tabular}{c | p{20 pt}<{\centering} p{15 pt}<{\centering} p{15 pt}<{\centering} p{15 pt}<{\centering} p{15 pt}<{\centering}}
\hline
  Method & $\text{mAP}^{\text{bbox}}$ & mIoU & $\text{AP}_{\text{vol}}^{\text{p}}$ & $\text{AP}_{\text{50}}^{\text{p}}$ & $\text{PCP}_{\text{50}}$ \\
 \hline
 Parsing R-CNN \cite{yang2019parsing} &68.7 & 56.3 & 53.9 & 63.7 & 60.1 \\
 Parsing R-CNN(w/ GN)* & -- & 56.2 & 54.2 & 64.6 & 60.9 \\
 \hline
 Parsing R-CNN(w/ FCOS) & 72.0 & 56.3 & 54.5 & 65.2 & 63.5 \\
  AIParsing(w/o refine) & \textbf{73.3} & \textbf{57.9} & \textbf{56.1} & \textbf{68.3} & \textbf{66.3} \\
\hline

\end{tabular}

}

\end{table}

\begin{table}
\centering
 \caption{Different multi-scale context modules on CIHP validation set. }
 \label{table:multiscale}
\footnotesize{
\begin{tabular}{c | c | c c c c }
\hline
  Multi-scale Context & mIoU & $\text{AP}_{\text{vol}}^{\text{p}}$ & $\text{AP}_{\text{50}}^{\text{p}}$ & $\text{PCP}_{\text{50}}$ \\
 \hline
 PSP \cite{zhao2017pyramid} &  53.1 & 52.0  & 59.1   &  57.7 \\
 ASPP \cite{chen2018encoder} & 53.2 &  52.4 & 59.8 & 58.5 \\
 PGEC \cite{zhang2020human} & \textbf{54.0}  &  \textbf{52.6} & \textbf{60.5}  & \textbf{59.6}  \\

\hline
\end{tabular}
}
\end{table}

\begin{table}[h]

\centering
\footnotesize{
\caption{Inference time comparison on CIHP validation set.}
 \label{table:runtime}
\begin{tabular}{c | c | c | c | c }
\hline
  Method & mIoU & Backbone & Params (M) &  FPS \\
 \hline
 PGN \cite{gong2018instance} & 54.4 & ResNet-101 & 629.2 & 4.2 \\
 Graphonomy \cite{gong2019graphonomy} & 55.5 & Xception & 156.9 & 25.0 \\
 GPM \cite{he2019grapy} & 56.2 & Xception & 176.0 & 16.6 \\
 \hline
 Parsing R-CNN \cite{yang2019parsing}  & 56.3 & ResNet-50 & 54.3 & 7.4 \\
 RP R-CNN \cite{yang2020eccv} & 58.2 & ResNet-50 & 58.4 & 5.0  \\
 AIParsing & 59.7 & ResNet-50 & 50.2 &  8.9 \\
 AIParsing-light & 56.7& ResNet-50 & 28.5 & 13.3 \\
\hline
\end{tabular}
}
\end{table}

\begin{figure}[t]
  \centering
  \includegraphics[width=0.9\linewidth]{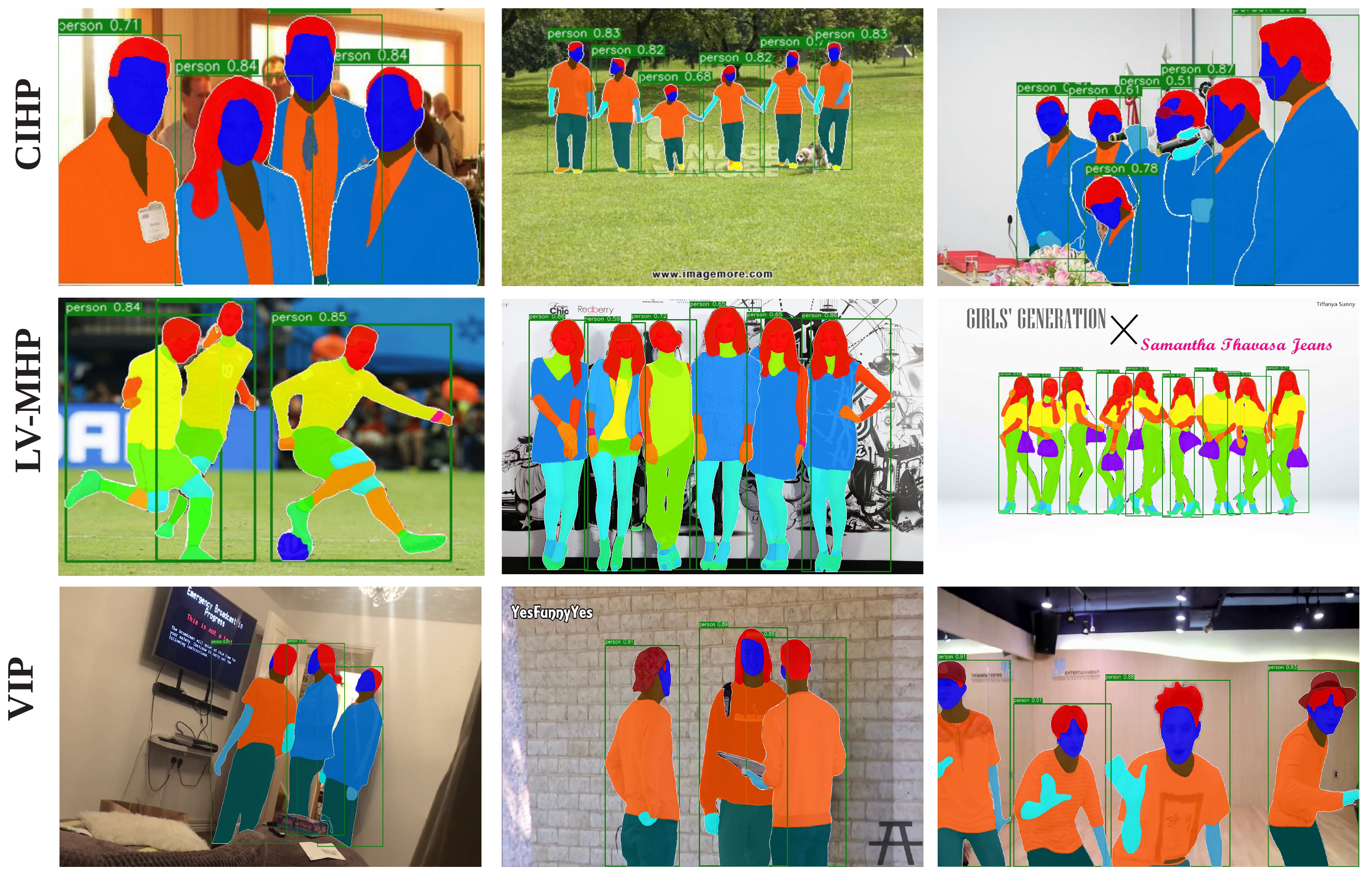}
  \caption{Visualization results on CIHP \cite{gong2018instance}, LV-MHP-v2.0 \cite{zhao2018understanding} and VIP \cite{zhou2018adaptive} val sets. The boxes with green color denote the human detection results.
   }
  \label{fig:final}
\end{figure}

\noindent\textbf{Different Multi-scale Context:} We choose PSP \cite{zhao2017pyramid} and ASPP \cite{chen2018encoder} for multi-scale context comparison, as shown in Table \ref{table:multiscale}. Here we just choose the FPN and human-part context encoding part for comparison. The differences are that we use PSP or ASPP to replace the PGEC module. We can find that the PGEC module is superior than PSP and ASPP modules both on global-level (mIoU) and instance-level ($\text{AP}_{\text{vol}}^{\text{p}}$, $\text{AP}_{\text{50}}^{\text{p}}$, $\text{PCP}_{\text{50}}$) metric scores. The good performance shows that the PGEC module is more suitable than PSP and ASPP for the human parsing task.

\noindent\textbf{Runtime Analysis:} We choose the CIHP validation set for runtime analysis, and we compare with three bottom-up based methods (PGN \cite{gong2018instance}, Graphonomy \cite{gong2019graphonomy}, GPM \cite{he2019grapy}) and two one-stage top-down instance-level human parsing methods (Parsing R-CNN \cite{yang2019parsing} and RP R-CNN \cite{yang2020eccv}). For a fair comparison, all methods are conducted on a single NVIDIA GeForce RTX 2080 Ti GPU with 11GB memory, the comparison results are shown in Table \ref{table:runtime}. The bottom-up based Graphonomy model can obtain the best FPS score compared with other methods. However, the Params of Graphonomy equipped with Xception backbone is larger than the proposed AIParsing-lite with ResNet-50 backbone. Meanwhile, the mIoU score of AIParsing-lite is superior to the Graphonomy method.
Our proposed AIParsing can obtain a better mIoU score and inference faster than other one-stage top-down parsing methods. The advantage benefits from the anchor-free FCN-like design. To further explore a faster
inference model, we designed a lighter AIParsing model to accelerate, which is named AIParsing-light. Specifically, except for the backbone part, the channels of all conv layers are reduced by half.
The number of serial conv layers for predicting branches in the detection head and edge-guided parsing head is also reduced by half. In addition, the input image scales are randomly sampled from [580, 600] pixels, and the images are resized to 600 pixels on the short side and 1,000 pixels or less on the long side. Other settings are the same as the standard AIParsing model. As shown in Table \ref{table:runtime}, although the parsing performance of AIParsing-light is worse than other methods, the AIParsing-light model owns fewer parameters and faster inference time than other comparison methods where it can run at 13.3 FPS.

\subsection{Qualitative Results}
In Fig. \ref{fig:three}, we visualize some qualitative comparison results with state-of-the-art one-stage top-down human parsing methods, all methods employ the same ResNet-50 backbone for fair comparison on CIHP validation set.
As shown in Fig. \ref{fig:three}, we can see that the proposed AIParsing can obtain better parsing results of adjacent human parts and more accurate instance regions than Parsing R-CNN \cite{yang2019parsing} and RP R-CNN \cite{yang2020eccv}. The human parsing results are usually influenced by the human detection results, if we predict accurate regions of the human instance, better parsing results will be obtained. As shown in Fig. \ref{fig:three} (a), (c), (f), our AIParsing provides relatively complete regions (denoted as quadrilateral regions) than Parsing R-CNN and RP R-CNN, especially in some challenging scenes, such as occlusion or appearance approximation. In Fig. \ref{fig:three} (f), the arm regions of two human instances are hard to distinguish. Benefit from taking the edge information into account, the model can distinguish two instance regions well.
In Fig. \ref{fig:three} (b), the proposed AIParsing predicts a more accurate scarf region than other ones (refer to the white elliptic region).
The proposed AIParsing predicts high-quality parsing results for the adjacent human part region, as shown in Fig. \ref{fig:three} (b) and (d).
These results reveal that our proposed method has the superiority of dealing with cases in the real-world scene.

Besides, we also show more visualization results obtained by our proposed AIParsing. Specifically, the qualitative results of AIParsing with a ResNeXt-101-32x8d backbone are shown in Fig. \ref{fig:final}, which are obtained from the validation sets of CIHP \cite{gong2018instance}, LV-MHP-V2.0 \cite{zhao2018understanding} and VIP \cite{zhou2018adaptive}, respectively. The experimental results demonstrate the effectiveness of the proposed AIParsing in various scenes, such as crowded scenes, overlapping instances, complex human poses, and so on. More visualization results can be found in the supplementary file.

\begin{figure}[h]
  \centering
  \includegraphics[width=0.8\linewidth]{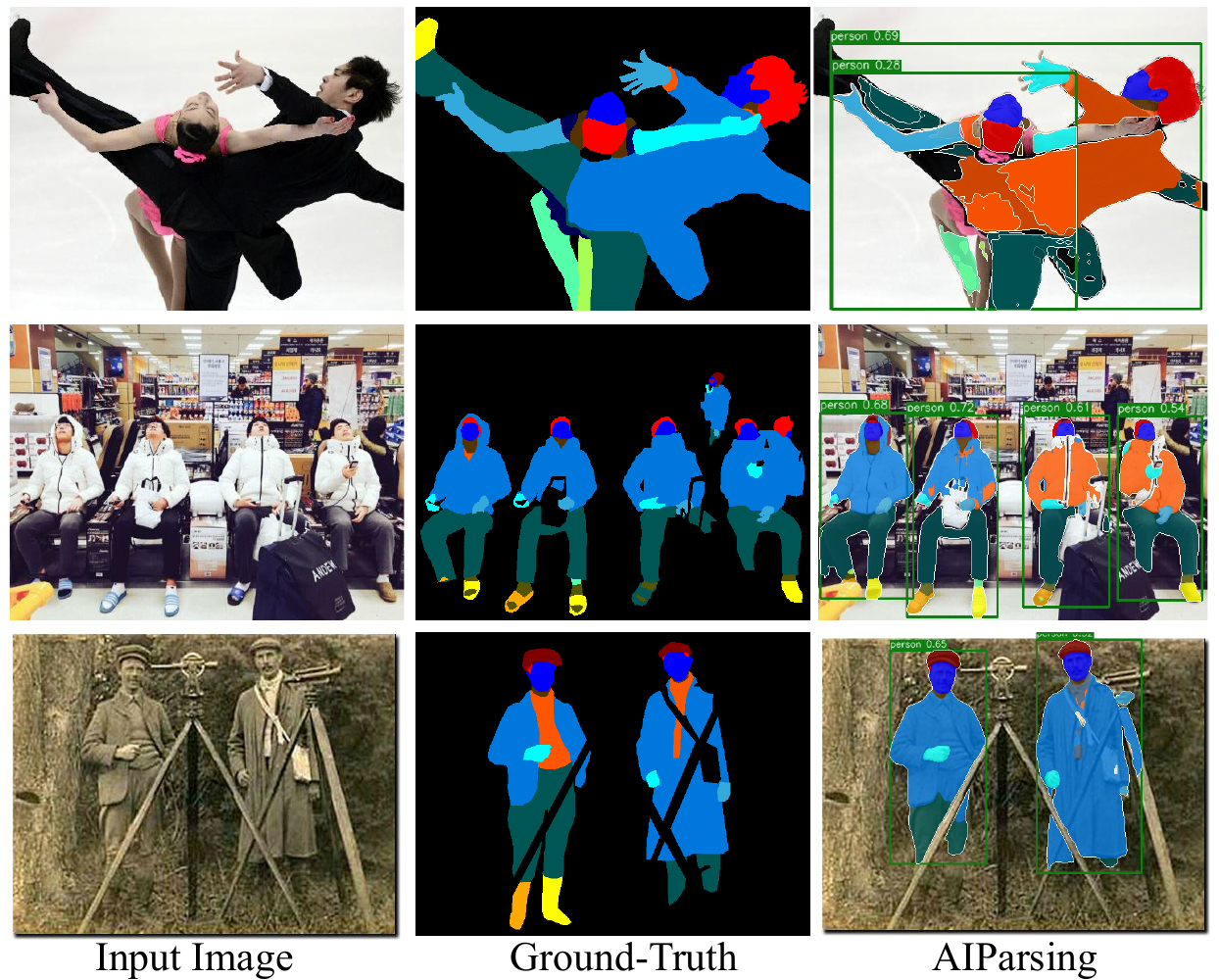}
  \caption{
    Some failure cases obtained from CIHP dataset.
   }
  \label{fig:error}
\end{figure}

\textbf{Failure Cases:} In some cases, the proposed AIParsing fails to obtain ideal parsing results, as shown in Fig. \ref{fig:error}, for example, highly overlapped instances,
close human classes, and the human instance region is close to the background. For the first row of Fig. \ref{fig:error}, these two human instances are highly overlapped, and the parsing result fails to accurately locate the instance region. For the second row of
Fig. \ref{fig:error}, four human instances are with the same coats, but the parsing results are not consistent in these regions.
For the third row of Fig. \ref{fig:error}, the background color is similar to the human instances, the anchor-free human detector fails to locate the whole instance region.
Thus, in the future, the multiple human parsers should improve the ability of dealing with highly overlapped instances and similar backgrounds, and consider the relationships among different human instances to avoid the inconsistent result in the same image.

\section{Conclusion}\label{conlusion}

In this paper, we propose a novel Anchor-free Instance-level human Parsing network (AIParsing), which fuses an anchor-free detector head and an edge-guided parsing head into a unified framework. The anchor-free detector has been certificated that it owns better ability than the anchor-based detector. In addition, edge information also provides help to distinguish adjacent human parts and overlapping instances. A refinement head is employed to enhance the parsing result through box-level mIoU-score and part-level mIoU-loss.
The proposed AIParsing outperforms current state-of-the-art methods especially on instance-level parsing metrics on two multiple human parsing datasets (\textit{i.e.}, CIHP and LV-MHP-v2.0) and one video human parsing dataset (\textit{i.e.}, VIP). We hope the FCN-like AIParsing model can be extended to solve other instance-level dense prediction tasks in a more detailed per-pixel fashion. Human structure is an important clue to help to parse the human parts well, which has been well certificated in human parsing tasks \cite{li2022deep, wang2019learning}. The core idea is to parse the human on multiple levels.  Thus, how to design a reasonable network taking the human structure into account needs further discussion in the future.
In addition, how to utilize instance-level human parsing technology solving the problems of human-centric video analysis \cite{lin2020human} especially in complex events is worth studying.


\ifCLASSOPTIONcaptionsoff
  \newpage
\fi



%
\footnotesize
\bibliographystyle{unsrt}
\bibliography{egbib}

\end{document}